\documentclass[10pt,twocolumn,letterpaper]{article}

\usepackage{cvpr}              

\usepackage[hyphens]{url}
\usepackage{graphicx}
\usepackage{amsmath}
\usepackage{amssymb}
\usepackage{mathtools}
\usepackage{booktabs}
\usepackage{multirow}
\usepackage{bm}
\usepackage{makecell}

\usepackage[pagebackref,breaklinks,colorlinks,hyperfootnotes=false]{hyperref}

\usepackage[capitalize]{cleveref}
\crefname{section}{Sec.}{Secs.}
\Crefname{section}{Section}{Sections}
\Crefname{table}{Table}{Tables}
\crefname{table}{Tab.}{Tabs.}

\graphicspath{{Figures/}}

\def\confName{CVPR}
\def\confYear{2027}

\makeatletter
\def\abstract
   {%
   \thispagestyle{empty}
   \centerline{\large\bf Abstract}%
   \vspace*{6pt}%
   \it%
   }

\makeatother

\makeatletter
\AtBeginDocument{%
  \renewcommand{\@makefntext}[1]{\parindent 1em\noindent #1}%
}
\makeatother

\begin{document}

\title{Multi-View Mixture-of-Experts with Vision-Language Reranking for Cross-View Object Geo-Localization}

\author{Xuyu Fan$^{1}$, Qi Ming$^{1,*}$, Zhu Han$^{1}$, Liuqian Wang$^{2}$, Si-Yuan Cao$^{3}$,\\
Xiaohan Zhang$^{3}$, Xudong Zhao$^{4}$, Mingjing Zhao$^{5}$, Yuhan Zhang$^{6}$\\
$^{1}$College of Computer Science, Beijing University of Technology,\\
$^{2}$Zhengzhou University, $^{3}$Zhejiang University, $^{4}$Beijing Institute of Technology,\\
$^{5}$Beijing Electronic Science and Technology Institute,\\
$^{6}$Intelligent Science \& Technology Academy of CASIC\\
{\tt\small fanxy6522@mails.jlu.edu.cn, chaser.ming@gmail.com}
}
\maketitle
\footnotetext{\hspace{1pc}*\ Corresponding author.}
\footnotetext{This work was completed during an internship and collaboration with Beijing University of Technology.}

\begin{abstract}
    Cross-view object geo-localization (CVOGL) locates a target in satellite imagery using drone or street-view queries. Existing methods train separate detectors for each viewpoint, leading to parameter redundancy and impeding cross-view knowledge sharing. Moreover, top-ranked satellite candidates are often visually similar, so visual appearance and categorical labels alone are insufficient to resolve such ambiguity. To address these, we propose MVLGeo, an efficient framework designed to unify multiple viewpoints and reduce model redundancy. First, we introduce environmental contextual text from the query view as cues to distinguish visually similar candidates via Vision-Language Reranking (VL-Rerank). Second, we design a multi-view Mixture-of-Experts architecture (MV-MoE) with a shared encoder and view-specific experts to reduce redundancy and promote knowledge sharing, while cross-view contrastive learning aligns their representations for consistency. Third, we introduce an adaptive elliptical prior (ESAM-Prior) as auxiliary positional encoding for anisotropic geometric perception. Extensive experiments on the CVOGL benchmarks confirm that MVLGeo, as a unified model for multiple query viewpoints, achieves state-of-the-art performance, demonstrating robustness to input degradation and generalization across viewpoints. Code and models will be available on GitHub to facilitate future work. 
\end{abstract}

\section{Introduction}
\begin{figure}[t]
    \centering
    \includegraphics[width=\linewidth]{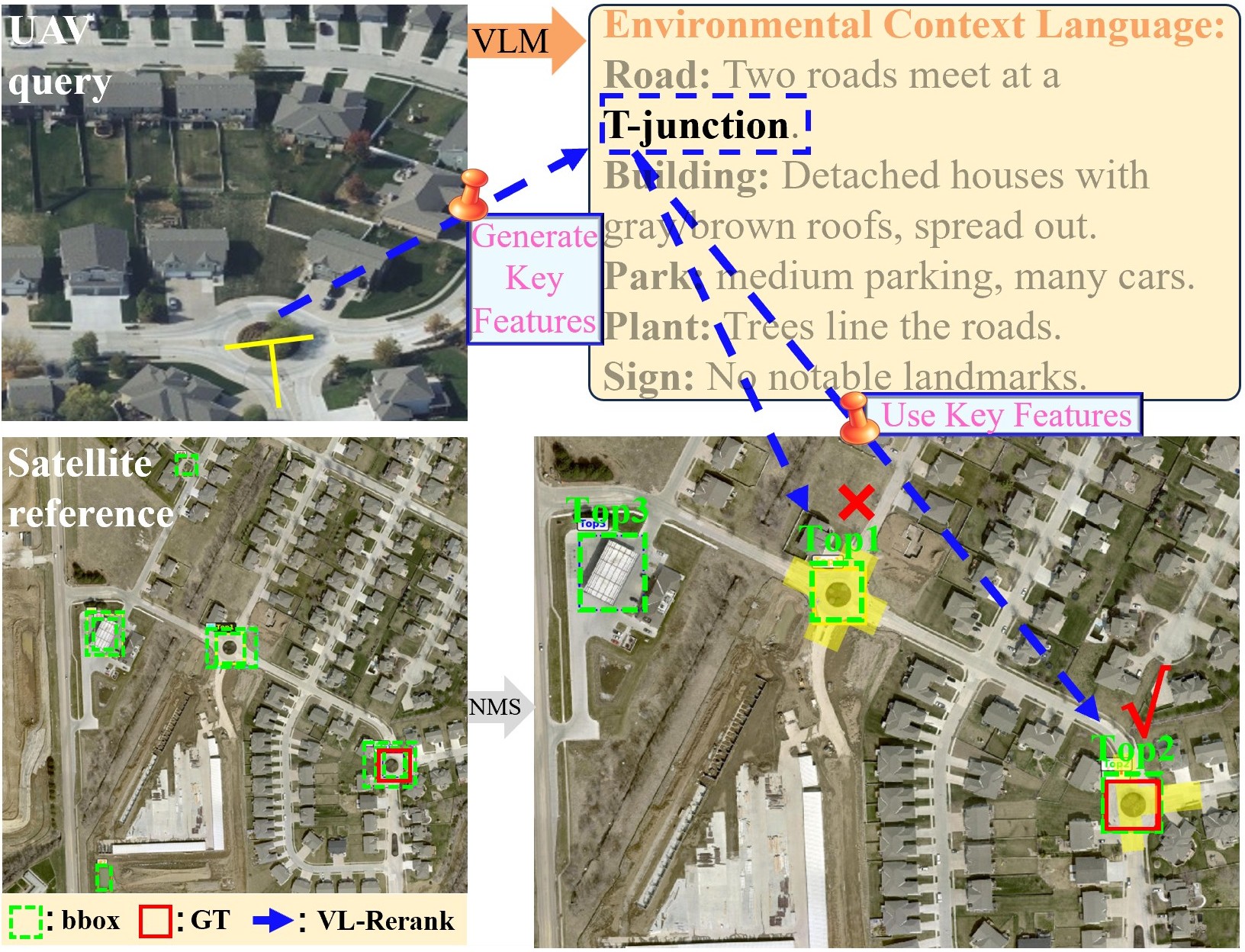}
    \caption{Illustration of the VL-Rerank framework. A VLM generates environmental context (e.g., road topology) from query images. These descriptions rerank satellite candidates, filtering out visually similar but contextually mismatched bounding boxes and identifying the correct target.}
    \label{fig:vl_rerank}
\end{figure}

Cross-view object geo-localization (CVOGL) aims to pinpoint a target in satellite imagery using UAV or SVI (street-view imagery) queries, serving as a fundamental capability for UAV navigation in GPS-denied environments, real-time urban monitoring, and post-disaster situation assessment. To bridge the geometric gap between viewpoints, existing approaches address this task through two main paradigms. Image-level methods~\cite{hu2018cvmnet,yang2021l2ltr,zhu2022transgeo} retrieve the most similar satellite patch but only provide coarse location estimates. Recently, DetGeo~\cite{sun2023detgeo} pioneered object-level CVOGL with click-prompted cross-view fusion, enabling direct bounding box regression from UAV to satellite imagery. Subsequent works further improved positional encoding~\cite{huang2025ocgnet,ghgeo2025}, detection architectures~\cite{afgeo2025,vageo2025}, and output refinement~\cite{zhang2025trogeo}. More recent efforts have explored rotated bounding boxes for orientation awareness~\cite{fu2026osgeo} and single-model localization that adapts to varying orientations and FoVs within street-view queries~\cite{chen2026singeo}.

\begin{figure}[t]
    \centering
    \includegraphics[width=\linewidth]{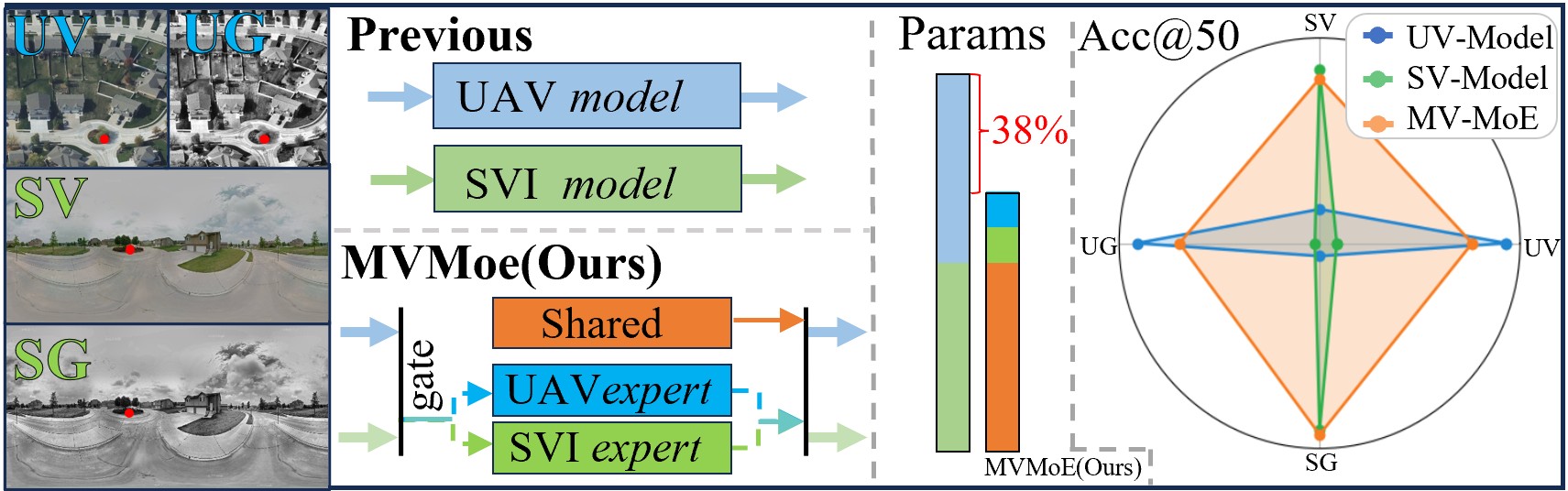}
\caption{Comparison of the proposed MV-MoE and independent models. MV-MoE shares one encoder and two lightweight experts (UAV, SVI) for four input variants (UAV Vis, UAV Gray, SVI Vis, SVI Gray), reducing parameters by 38\% while maintaining Acc@50 (\%).}
    \label{fig:mvmoe_arch}
\end{figure}

Despite these advances, current object-level CVOGL methods face fundamental limitations: (1) \textbf{Candidate ambiguity.} We observe that after detection and non-maximum suppression, the top-ranked satellite candidates are often visually indistinguishable from each other. Their confidence scores cluster within a narrow margin. As Fig.~\ref{fig:vl_rerank} shows, the flowerbeds of roundabouts appear as nearly identical green circular patches in satellite imagery. Category-level text labels~\cite{wang2024geotext,zhang2026mopt} cannot resolve this ambiguity, because they carry no spatial or topological context. This suggests that richer, spatially grounded textual cues could provide the missing disambiguation signal. (2) \textbf{Inherent viewpoint non-transferability.} Existing methods adopt a viewpoint-specific training paradigm, where a separate detector is trained for each viewpoint. Such a paradigm achieves satisfactory performance only on the trained viewpoints, but degrades drastically on unseen ones. Different configurations thus require different models, even though visual features across views are largely shared. DetGeo~\cite{sun2023detgeo}, for example (illustrated in Fig.~\ref{fig:mvmoe_arch}), deploys two separate models, each trained independently with its own parameters to handle a specific viewpoint. This per-viewpoint paradigm leads to near-linear growth in parameters and computational cost, making multi-domain deployment increasingly expensive and inflexible as the number of viewpoints grows. (3) \textbf{Insufficient input prior.} Most existing encodings are direction-agnostic. For elongated or irregular objects, isotropic priors spread activation into background regions, introducing noise that impairs precise localization. This problem is specific to object-level CVOGL, as in image-level cross-view geo-localization (CVGL), background context contributes to scene-level matching and is therefore benign, but CVOGL regresses a precise bounding box, so any prior activation that leaks into the background directly pulls the regression away from the target boundary. An anisotropic shape-aware prior would confine attention to the target region, yet such a prior is missing from current methods.

To tackle these challenges, we design a unified framework, MVLGeo. (1) We observe that environmental context, including road topology, building layout, and vegetation, constitutes an instance-level spatial fingerprint that remains discriminative when matching a query to its satellite counterpart. This motivates Vision-Language Reranking (VL-Rerank), where we apply prompt engineering to extract structured environmental descriptions from the query view, then fine-tune a VLM to compare and rerank candidates, providing a disambiguation signal beyond visual features and class labels. (2) Instead of training separate models for each viewpoint, we adopt the Mixture-of-Experts paradigm~\cite{liu2024tmlr-routers} and design our Multi-View Mixture-of-Experts (MV-MoE) for parameter sharing rather than capacity scaling, where a shared encoder captures cross-view generic features, while lightweight view-specific experts learn only minimal view-specific offsets. Hard routing by view identity replaces learned gating, and a cross-view contrastive loss~\cite{chen2020simclr} prevents expert divergence. (3) We note that SAM~\cite{kirillov2023sam} excels at isolating objects from backgrounds via point-prompted segmentation. We adapt this capability to CVOGL by fitting oriented ellipses to SAM masks, producing anisotropic Gaussian heatmaps (ESAM-Prior) that capture instance-specific shape and orientation.

Our contributions can be summarized as follows:
\begin{itemize}
\item We show that cross-view invariant environmental topology resolves candidate ambiguity where visual features and category labels fail. VL-Rerank introduces language modality through environmental textual descriptions, providing discriminative cues beyond visual appearance.
\item MVLGeo is a unified multi-view framework for CVOGL that explicitly shares parameters across viewpoints via Mixture-of-Experts. Our MV-MoE architecture replaces per-view detectors with a shared encoder and identity-initialized view-specific experts, reducing total parameters while maintaining competitive accuracy across all viewpoints.
\item We introduce an elliptical adaptive prior (ESAM-Prior) as a shape-aware positional encoding. By fitting an oriented ellipse to the target mask, this prior provides anisotropic geometric guidance for elongated and irregular targets.
\end{itemize}

\section{Related Work}

\begin{figure*}[!t]
    \centering
    \includegraphics[width=\textwidth]{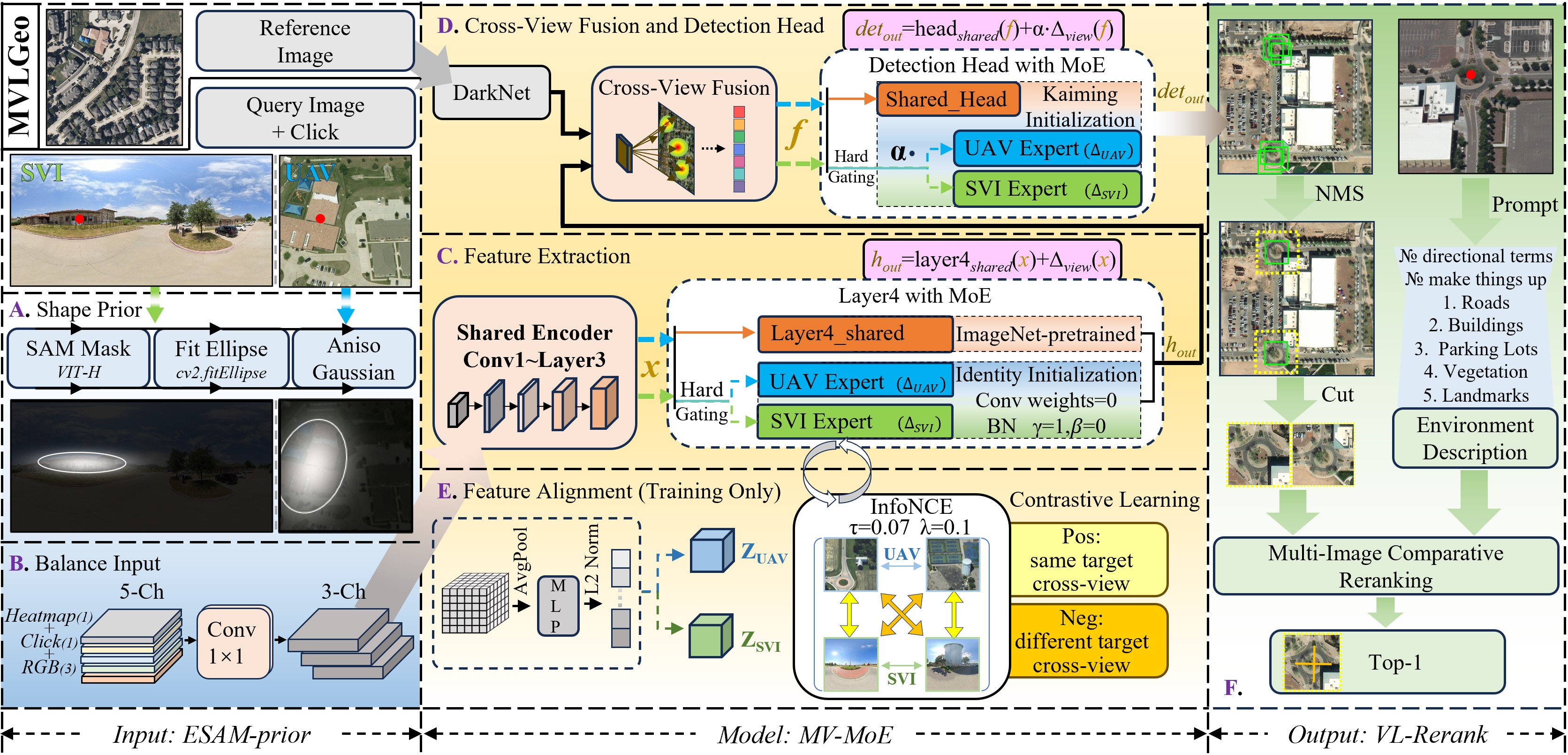}
    \caption{Schematic diagram of the MVLGeo network architecture. (A) Input-guided features are generated using SAM and shape priors. (B) Balancing heterogeneous input channels. (C--D) Using a Mixture-of-Experts (MoE) architecture with hard gating to perform cross-view feature extraction, fusion, and detection from UAV and SVI perspectives. (E) Aligning cross-view features during the training phase via InfoNCE contrastive learning. (F) Performing multi-image comparison and reranking by combining environmental semantic descriptions and image cropping to output the Top-1 matching result.}
    \label{fig:MVLGeo}
\end{figure*}

\textbf{Cross-View Object Geo-Localization.} DetGeo~\cite{sun2023detgeo} pioneered object-level CVOGL with click-prompted cross-view fusion, enabling direct bounding box regression from UAV or street-view queries to satellite imagery. Subsequent works improve positional encoding~\cite{huang2025ocgnet,ghgeo2025}, detection architectures~\cite{afgeo2025,vageo2025}, and output refinement~\cite{zhang2025trogeo}, with recent extensions to oriented boxes~\cite{fu2026osgeo}, multi-object scenarios~\cite{lv2026mogeo}, and single-stage frameworks~\cite{wang2026gageo}. These methods improve localization accuracy but remain viewpoint-specific and rely on isotropic positional priors, which are ill-suited for precise localization of elongated targets. SAM~\cite{kirillov2023sam} has been applied in CVOGL for output-side refinement (TROGeo~\cite{zhang2025trogeo}) and mask-driven encoding (EDGeo~\cite{hu2025edgeo}), but using SAM-derived anisotropic priors as input-level positional encoding remains unexplored.

\textbf{Mixture-of-Experts in Cross-View Geo-Localization.} Classic MoE~\cite{shazeer2017moe} scales model capacity through learned gating that sparsely activates experts, extended by GShard~\cite{lepikhin2021gshard} and Switch Transformer~\cite{fedus2022switch} to large-scale models. Subsequently, the MoE paradigm has continued to evolve, with Expert Choice~\cite{zhou2024expertchoice} inverting the routing direction, DeepSeekMoE~\cite{dai-etal-2024-deepseekmoe} introducing shared experts with fine-grained segmentation, and Soft MoE~\cite{puigcerver2024softmoe} replacing discrete routing with soft assignments, broadening MoE into a more general conditional computation framework. Liu~\cite{liu2024tmlr-routers} and Wong~\cite{wong2026calibrated} further formalize hard-routed MoE as routing each input to a single expert, providing a theoretical basis for our view-identity-based hard routing. In CVOGL, SMGeo~\cite{zhang2025smgeo} introduces grid-level sparse MoE for capacity expansion, and HiSymGeo~\cite{chen2026hisymgeo} employs query-gated multi-expert feature fusion. Concurrent efforts explore platform-level expert partitioning~\cite{li2025moe} and KV routing for cross-view alignment~\cite{liu2025learnable}. Most existing MoE applications in CVOGL adopt learned routing for capacity scaling, while unifying multiple query viewpoints within a single detection model remains largely unexplored.

Beyond MoE-based methods, SinGeo~\cite{chen2026singeo} targets orientation and FoV robustness within street-view queries through curriculum learning, while SkyPart~\cite{tran2026skypart} addresses visual degradation via invariant texture topology. These works reduce per-condition model replication but focus on robustness within a single viewpoint.

\textbf{Vision-Language Models for Geo-Localization.} CLIP~\cite{radford2021clip}, BLIP-2~\cite{li2023blip2}, and InternVL~\cite{chen2024internvl} have enabled cross-view spatial reasoning through vision-language alignment. Existing text-guided methods fall into two categories. Category-level supervision associates object classes with visual features (GeoText~\cite{wang2024geotext}, MoPT~\cite{zhang2026mopt}), but class labels lack the spatial and topological detail needed to distinguish visually similar candidates. Scene-level methods leverage hierarchical descriptions for image matching (GeoMatch~\cite{wu2026geomatch}), semantic-anchored multi-view models (GeoBridge~\cite{song2026geobridge}), or VLM-based reranking (GeoVLM~\cite{dagda2025geovlm}, CLIP-UG~\cite{wu2025clipug}), while GeoX-Bench~\cite{zheng2026geoxbench} provides a benchmark for evaluating such methods. However, existing methods either rely on coarse category labels or operate at the scene level. Using VLM-based comparative reasoning with viewpoint-invariant environmental descriptions for object-level candidate disambiguation remains open.

\section{Methodology}

\subsection{Overview}
We build upon DetGeo~\cite{sun2023detgeo} as our baseline. MVLGeo improves the baseline from three perspectives: instance-adaptive anisotropic positional encoding for better geometric perception, a multi-view unified architecture that eliminates per-view parameter redundancy, and language-guided candidate disambiguation. As illustrated in Fig.~\ref{fig:MVLGeo} (stages A to F), these components form an end-to-end unified framework. Both SAM priors and VLM descriptions can be precomputed offline and cached, with online generation supported for new queries.

\subsection{VL-Rerank: Vision-Language Reranking}
\label{sec:vlm}

\begin{figure}[!t]
    \centering
    \includegraphics[width=1\linewidth]{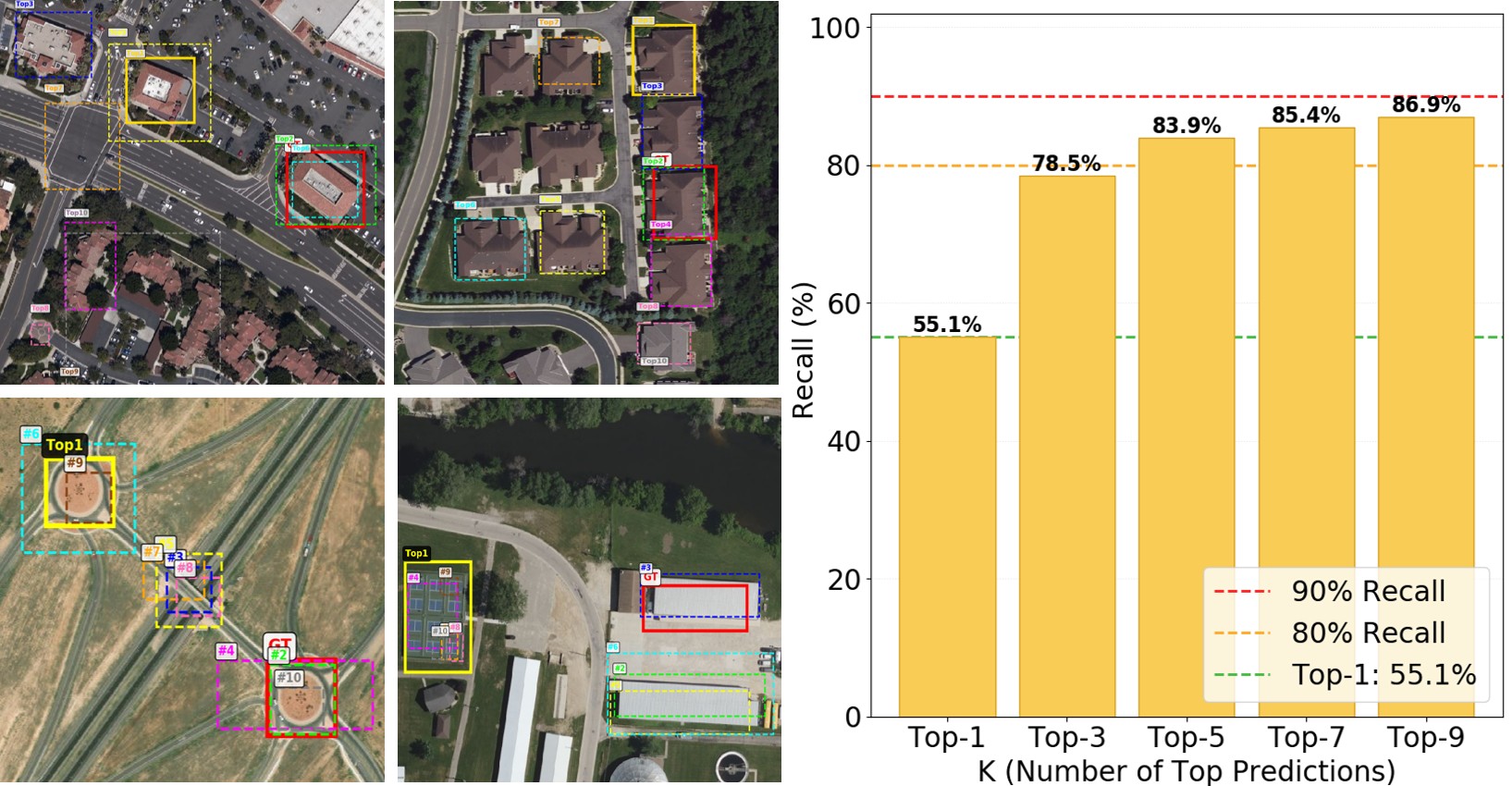}
    \caption{Recall@K reveals that the ground-truth box often appears among top candidates but is not ranked first.}
    \label{fig:Topk-recall}
\end{figure}

To resolve candidate ambiguity, we introduce a VLM-based reranking stage that leverages environmental context from the query view. As observed in Fig.~\ref{fig:Topk-recall}, the ground-truth box appears among the top-3 candidates in 78.3\% of cases but is ranked first only 55.1\%, confirming that visual appearance alone is insufficient for disambiguation. 

\subsubsection{Environmental Description Generation.}
To ensure viewpoint invariance, descriptions must avoid absolute directions (e.g., north of) or viewpoint-specific attributes, instead capturing stable spatial relations (e.g., adjacent to, surrounded by) that hold from any perspective. We design four complementary prompts for Qwen-VL-Chat, covering road topology, building layout, parking lots, vegetation, and landmarks, with a red circular marker overlaid at the click point to direct attention (used purely for VLM, not detection). The VLM outputs a few factual sentences, which are cached for reuse.

\subsubsection{VLM-Based Comparative Reranking.}
We fine-tune InternVL~\cite{chen2024internvl} for multi-image comparative reasoning. For each query, the top-$K$ satellite candidates are cropped, resized, color-bordered with numeric indices, and arranged into a composite image. This composite, paired with the environmental description, forms the multi-modal input. Training data is generated offline using the frozen DetGeo detector, and for each sample we composite the top-3 candidates and select the one with highest IoU to the ground truth as the target label. Data is formatted in the LLaMA-Factory chat template for next-token prediction fine-tuning. At inference, InternVL directly outputs the best candidate index, enabling end-to-end visual comparison without intermediate scoring functions.

\subsection{MV-MoE: Multi-View Mixture-of-Experts Architecture}
\label{sec:mv_moe}

DetGeo trains independent models per viewpoint. Extending to $V$ viewpoints would require $V$ full detectors. Yet the underlying task is shared. Only the statistical distribution differs, not the visual concepts. This suggests most parameters can be shared, with minimal view-specific adaptations.

The MV-MoE architecture (C--D) in Fig.~\ref{fig:MVLGeo} operationalizes this via a shared-residual decomposition. At each expert insertion point $\ell$, a view's function is factorized as:
\begin{equation}
f_{view}^{(\ell)}(x) = f_{shared}^{(\ell)}(x) + \Delta_{view}^{(\ell)}(x), \label{eq:decomp}
\end{equation}
where $f_{shared}^{(\ell)}$ captures generic capabilities jointly optimized across views, and $\Delta_{view}^{(\ell)}$ compensates for view-specific discrepancies. Each residual is initialized near zero, so the model learns only minimal offsets from a sharing-optimal state. Our MV-MoE adopts hard routing by view identity instead of learned gating. This eliminates router overhead and unstable dynamics, which is particularly beneficial for our parameter-sharing objective:
\begin{equation}
\min_{\theta_{shared}, \{\theta_{view}\}} \smashoperator[r]{\sum_{v \in \{\mathit{UAV}, \mathit{SVI}\}}}\! \mathcal{L}_{det}\!\left(f_{shared}(x_v) + \Delta_v(x_v),\, y_v\right).
\end{equation}
The complete training objective additionally incorporates cross-view contrastive alignment (Eq.~\ref{eq:total_loss}).

We insert experts at two positions. The first is ResNet18 layer4 (C), which encodes high-level semantic features where cross-view discrepancies manifest while early layers (conv1 to layer3) encode viewpoint-invariant edges and textures, making them unsuitable for expert insertion:
\begin{equation}
h_{out} = \text{layer4}_{shared}(x) + \Delta_{view}(x).
\end{equation}
Concretely, the view-specific expert $\Delta_{view}$ at layer4 consists of two $3{\times}3$ convolutional layers with batch normalization and ReLU, matching the residual block design of ResNet's layer4. The final convolution weights are zero-initialized and the final BN is initialized with $\gamma{=}1,\beta{=}0$, ensuring $\Delta_{view}(x) = 0$ at initialization and thus identity mapping from the pretrained backbone. The detection head expert (D) consists of a single $1{\times}1$ convolution followed by batch normalization, reducing from 512 channels to 54, the YOLOv3 output dimension, with weights drawn from a zero-mean Gaussian ($\sigma{=}10^{-4}$) to keep its initial contribution near zero. It operates on the fused feature $\mathbf{f}$:
\begin{equation}
det_{out} = \text{head}_{shared}(\mathbf{f}) + \alpha \cdot \Delta_{view}(\mathbf{f}),
\end{equation}
where $\alpha$ is a learnable scalar initialized to 1.0. Since the shared head is randomly initialized from scratch, the head expert uses Kaiming initialization with a small-weight final projection, keeping its initial contribution near zero. This design lets the view-specific expert adapt to viewpoint-specific textures (e.g., roof contours vs. facades) while the head expert adjusts detection preferences.

\subsubsection{Contrastive Feature Alignment.}

The shared-residual decomposition (Eq.~\ref{eq:decomp}) ensures each expert learns only offsets, but without constraint, UAV and SVI experts could drift into incompatible feature spaces. The gradients of their detection losses on the shared encoder exhibit conflicting directions, causing oscillation rather than convergence.

To address this, we apply a symmetric InfoNCE contrastive loss (E) in Fig.~\ref{fig:MVLGeo} that pulls corresponding cross-view features together while pushing non-matching pairs apart:
\begin{equation}
\mathcal{L}_{CL}\! = \!\frac{1}{2}\bigl[\text{InfoNCE}(z_{\mathit{UAV}}, z_{\mathit{SVI}}) \!+ \text{InfoNCE}(z_{\mathit{SVI}}, z_{\mathit{UAV}})\bigr].
\end{equation}
Here, $z_{view}$ is a normalized 128-dimensional embedding from a 2-layer MLP projection head (following SimCLR~\cite{chen2020simclr}). For a batch of $B$ aligned UAV-SVI pairs, positives are $(z_U^i, z_S^i)$ and negatives are $(z_U^i, z_S^j), i \neq j$. Gradient flow from $\mathcal{L}_{CL}$ is blocked from the detection head experts ($\alpha \cdot \Delta$) to preserve detection-level specialization, while updating the backbone and shared parameters to ensure well-aligned updates. The overall objective balances alignment and specialization:
\begin{equation}
\mathcal{L}_{total} = 0.5(\mathcal{L}_{det}^{\mathit{UAV}} + \mathcal{L}_{det}^{\mathit{SVI}}) + \lambda \cdot \mathcal{L}_{CL}. \label{eq:total_loss}
\end{equation}

\subsection{ESAM-Prior: SAM Adaptive Elliptical Prior}
\label{sec:sam}

\begin{figure}[!t]
    \centering
    \includegraphics[width=1\linewidth]{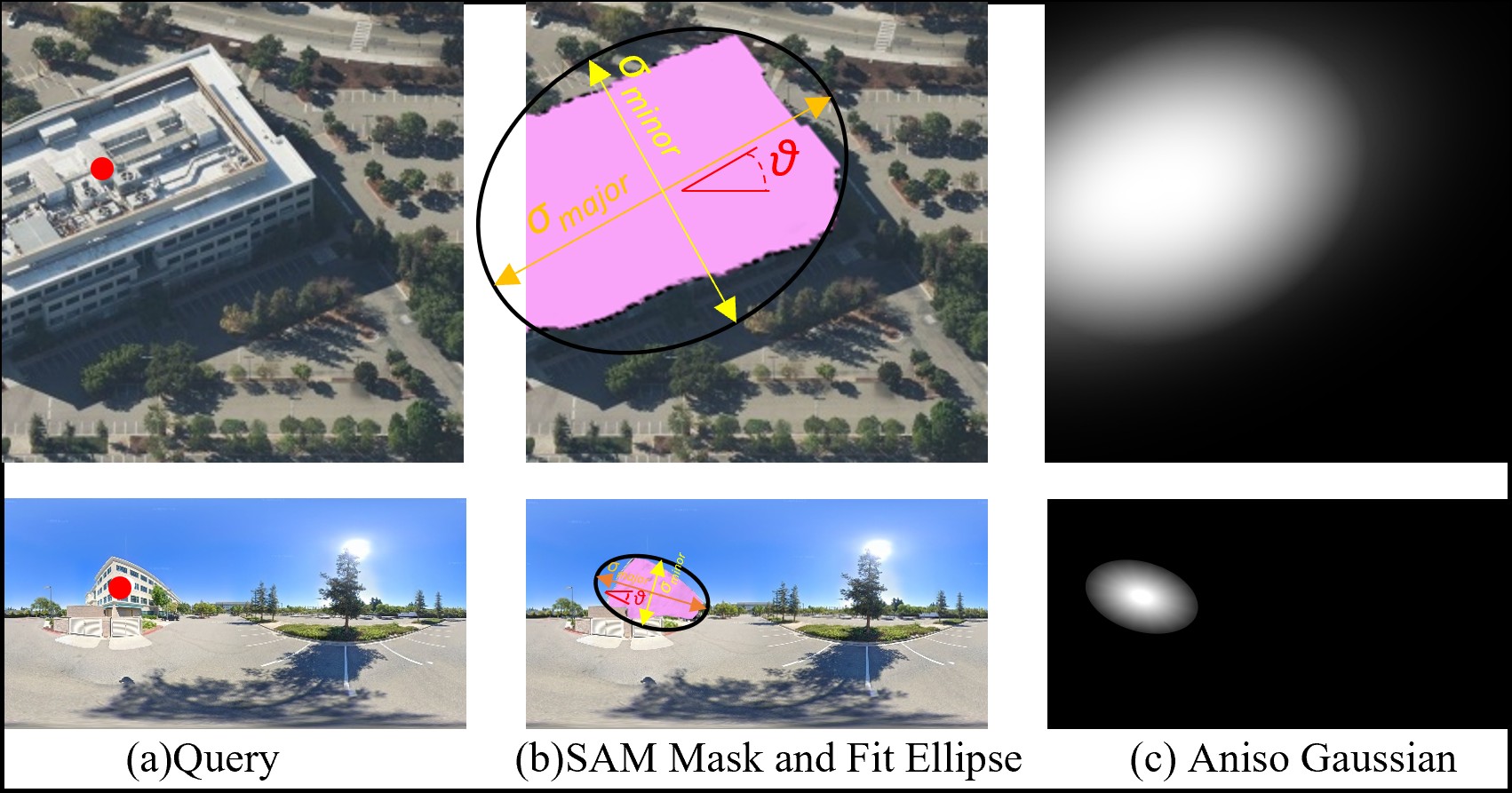}
    \caption{Shape prior generation pipeline. (a) Query image, (b) SAM mask and fitted ellipse, (c) Anisotropic Gaussian heatmap.}
    \label{fig:ESAM-Prior}
\end{figure}

We propose an instance-adaptive anisotropic prior that captures the target's shape and orientation at runtime, confining the positional signal to the target's geometric footprint rather than spreading into background regions.

\subsubsection{SAM-Guided Ellipse Fitting.}

Given the query image $I_{query}$ and click $(c_x, c_y)$, we first obtain a segmentation mask via SAM~\cite{kirillov2023sam} (ViT-H, point prompt). An oriented ellipse is then fitted to this mask to extract three parameters: major axis $\sigma_{major}$, minor axis $\sigma_{minor}$, and orientation $\theta$. Unlike fixed isotropic Gaussians that treat all targets identically, these parameters naturally adapt to each instance. Compact vehicles produce near-circular ellipses ($\sigma_{major}{\approx}\sigma_{minor}$), while elongated structures produce highly anisotropic ones ($\sigma_{major}{\gg}\sigma_{minor}$) with explicit directional alignment.

\subsubsection{Dual-Map Input Encoding.}

The extracted parameters are rendered as an anisotropic Gaussian heatmap centered at the click point (see Fig.~\ref{fig:ESAM-Prior})

\begin{equation}
P_{ellipse}(x, y) = \exp\!\left(-\frac{1}{2}\!\left[\left(\frac{x_{rot}}{\sigma_{major}}\right)^{\!2} \!+ \left(\frac{y_{rot}}{\sigma_{minor}}\right)^{\!2}\right]\right),
\end{equation}

where $(x_{rot}, y_{rot}) = \mathbf{R}(\theta)(x-c_x, y-c_y)$. Unlike isotropic priors that spread equally in all directions, this heatmap concentrates activation along the object's major axis and decays rapidly along the minor axis, confining the positional signal to the target's actual geometric footprint.

When SAM segmentation is unreliable, we fall back to a fixed circular Gaussian prior centered at the click point. Alongside the elliptical heatmap, we retain a binary click map to provide a precise center anchor. The two maps are concatenated with the RGB channels and compressed via a $1{\times}1$ convolution before entering the shared encoder, allowing the network to jointly leverage visual appearance, positional anchor, and shape prior.

\section{Experiments}

\begin{table*}[!t]
\centering
\small
\setlength{\tabcolsep}{3pt}
\begin{tabular}{c|cc|cc|cc|cc|c}
\toprule
 & \multicolumn{4}{c}{Ground{}$\rightarrow${}Satellite} & \multicolumn{4}{c|}{Drone{}$\rightarrow${}Satellite} & \\
\cmidrule(lr){2-5}\cmidrule(lr){6-9}
Method & \multicolumn{2}{c}{Validation} & \multicolumn{2}{c}{Test} & \multicolumn{2}{c}{Validation} & \multicolumn{2}{c|}{Test} & Models \\
\cmidrule(lr){2-3}\cmidrule(lr){4-5}\cmidrule(lr){6-7}\cmidrule(lr){8-9}
 & Acc@25\% & Acc@50\% & Acc@25\% & Acc@50\% & Acc@25\% & Acc@50\% & Acc@25\% & Acc@50\% & \\
\midrule
CVM-Net~\cite{hu2018cvmnet} & 5.09 & 0.87 & 4.73 & 0.51 & 20.04 & 3.47 & 20.14 & 3.29 & 2 \\
RK-Net~\cite{lin2022rknet} & 8.67 & 0.98 & 7.40 & 0.82 & 19.94 & 3.03 & 19.22 & 2.67 & 2 \\
L2LTR~\cite{yang2021l2ltr} & 12.24 & 1.84 & 10.69 & 2.16 & 38.68 & 5.96 & 38.95 & 6.27 & 2 \\
Polar-SAFA~\cite{shi2019polarsafa} & 19.18 & 2.71 & 20.66 & 3.19 & 36.19 & 6.39 & 37.41 & 6.58 & 2 \\
TransGeo~\cite{zhu2022transgeo} & 21.67 & 3.25 & 21.17 & 2.88 & 34.78 & 5.42 & 35.05 & 6.47 & 2 \\
SAFA~\cite{shi2019safa} & 20.59 & 3.25 & 22.20 & 3.08 & 36.19 & 6.39 & 37.41 & 6.58 & 2 \\
GeoDTR+~\cite{zhang2024geodtr} & 14.08 & 1.95 & 14.19 & 5.14 & 15.71 & 3.68 & 16.03 & 4.73 & 2 \\
\midrule
DetGeo~\cite{sun2023detgeo} & 46.70 & 43.99 & 45.43 & 42.24 & 59.81 & 55.15 & 61.97 & 57.66 & 2 \\
VAGeo~\cite{vageo2025} & 47.56 & 44.42 & 48.21 & 45.22 & 64.25 & 59.59 & 66.19 & 61.87 & 2 \\
OCGNet~\cite{huang2025ocgnet} & \textbf{48.54} & 44.20 & \textbf{51.49} & \textbf{47.69} & 66.52 & 61.86 & 68.35 & 63.93 & 2 \\
\midrule
\textbf{MVLGeo (Ours)} & 48.32 & \textbf{45.18} & 50.05 & 47.24 & \textbf{69.09} & \textbf{63.11} & \textbf{73.72} & \textbf{66.04} & \textbf{1} \\
\bottomrule
\end{tabular}
\caption{Comparison of Acc@25/50 (\%) and number of models on the CVOGL dataset. Bold denotes the best results.}
\label{tab:sota}
\end{table*}

\subsection{Dataset and Evaluation Metrics}

\textbf{Dataset.} CVOGL~\cite{sun2023detgeo} is a large-scale cross-view object geo-localization dataset. It supports two tasks: \textit{Drone$\rightarrow$Satellite} using UAV images as queries, and \textit{Street-View$\rightarrow$Satellite} (also referred to as \textit{Ground$\rightarrow$Satellite}) using street-view images as queries. The UAV query images are 256$\times$256, the street-view query images are 256$\times$512, and the satellite reference images are 1024$\times$1024. Each task is split into 4,343 training, 923 validation, and 973 test pairs.

\textbf{Evaluation Metrics.} Following DetGeo~\cite{sun2023detgeo}, we adopt Acc@50 (\%) and Acc@25 (\%) as evaluation metrics, where Acc@K measures the proportion of queries whose predicted bounding box achieves an IoU greater than $K$ with the ground truth. We report results on both validation and test sets. In addition to localization accuracy, we report the total number of model parameters to assess parameter efficiency. Unlike prior methods that train two separate detectors, one for each viewpoint, our framework uses a single unified model for both UAV and SVI, enabling a direct comparison of total parameter cost. We further introduce the average Acc@50 (\%) across different modalities and viewpoints as a new metric to evaluate cross-modal and cross-view generalization.

\subsection{Implementation Details}

All experiments are conducted on two NVIDIA RTX 4090 GPUs. We adopt the DetGeo~\cite{sun2023detgeo} framework as our base architecture. In the VL-Rerank, environmental descriptions are generated by Qwen-VL-Chat with structured prompts, and InternVL is fine-tuned as a multi-image comparative reranker using the LLaMA-Factory chat template. In the MV-MoE, we adopt ResNet18 pretrained on ImageNet-1K as the shared encoder, with view-specific experts inserted at the encoder and detection head. The model is trained with RMSProp for 25 epochs with a batch size of 12, alternating UAV and SVI samples per step. The learning rate is initialized at $1\times10^{-4}$ and decayed by a factor of 10 every 10 epochs. The contrastive learning weight $\lambda$ is set to 0.1. In the ESAM-Prior, we apply SAM (ViT-H) to generate instance-level segmentation masks, which are fitted with oriented ellipses to produce the anisotropic Gaussian prior.

\subsection{Comparison with State-of-the-Art Methods}

Tab.~\ref{tab:sota} compares MVLGeo with existing CVOGL methods. Results of prior works are taken from their original papers. Our method achieves state-of-the-art performance on the Drone$\rightarrow$Satellite (UAV) task and competitive accuracy on Ground$\rightarrow$Satellite (street-view, SVI). More importantly, this is accomplished with only a single unified model for both viewpoints, whereas all prior methods require training and deploying two independent detectors, one per viewpoint.

\begin{table}[!t]
\centering
\small
\setlength{\tabcolsep}{8pt}
\begin{tabular}{lccc}
\toprule
Method & Params & Models & Total \\
\midrule
DetGeo & 73.8M & 2 & 147.6M \\
OCGNet & 74.8M & 2 & 149.6M \\
\textbf{MVLGeo (Ours)} & 91.0M & \textbf{1} & \textbf{91.0M} \\
\bottomrule
\end{tabular}
\caption{Comparison of model parameters and deployment costs between existing dual-detector methods and our single unified model.}\label{tab:params}
\end{table}

To our knowledge, MVLGeo is the first framework to unify two distinct viewpoints (UAV and SVI) within a single detection model for object-level CVOGL. This architectural advantage is reflected in parameter efficiency and model count. As shown in Tab.~\ref{tab:params}, prior methods duplicate the entire detection, requiring twice the parameters of a single model for two viewpoints. MVLGeo unifies both tasks within a single model, reducing total model size by 38\%. This unified design confirms that the vast majority of visual knowledge is shared across viewpoints, with only a small fraction of parameters requiring view-specific adaptation. This paradigm opens a new direction for scalable deployment across arbitrary viewpoints without linear parameter growth, simplifying deployment and reducing memory footprint.

\subsection{Ablation Studies}
\label{sec:ablation}

\subsubsection{Component-wise Ablations.}
\begin{figure}[!t]
    \centering
    \includegraphics[width=1\linewidth]{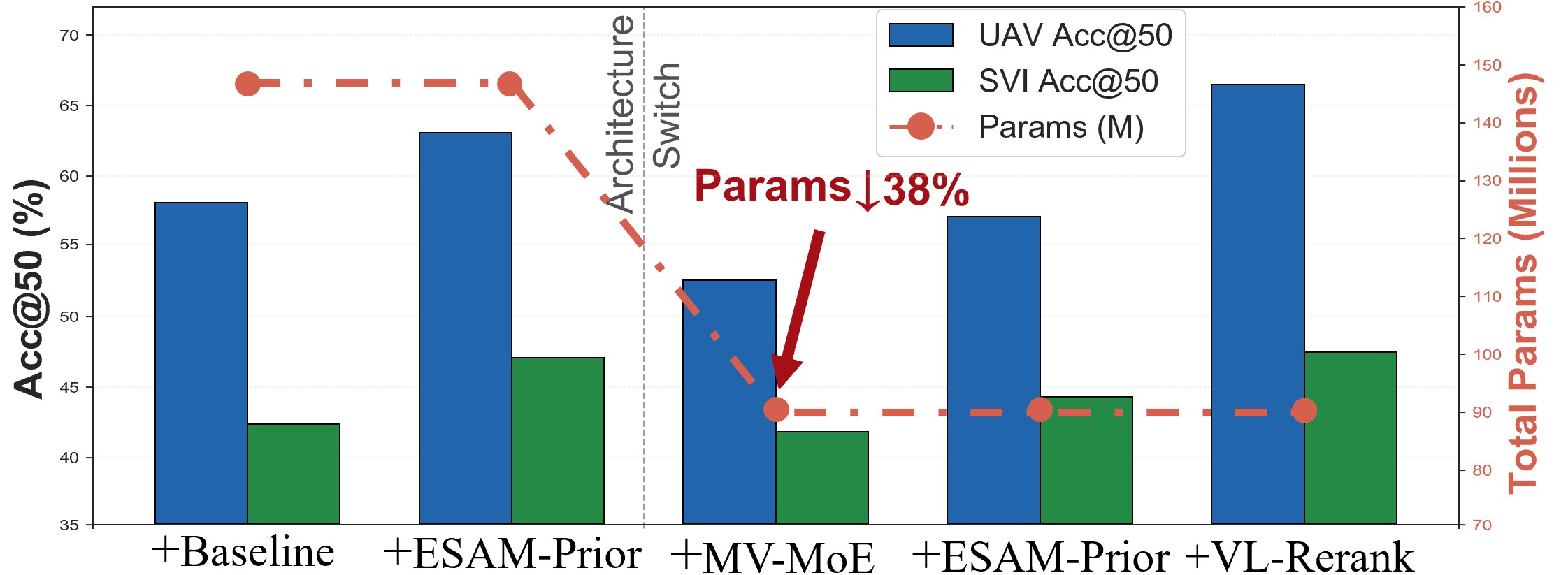}
    \caption{Contributions of each component to the UAV and SVI test sets, and corresponding parameter relationships.}
    \label{fig:Ablation_MVLGeo}
\end{figure}

To evaluate the MVLGeo framework, we conduct ablation studies on each component and perform multiple runs. The experimental results are summarized in Tab.~\ref{tab:modular} and Fig.~\ref{fig:Ablation_MVLGeo}. First, applying ESAM-Prior to the baseline~\cite{sun2023detgeo} yields a 4.9\% gain on UAV and 4.5\% on SVI over the baseline. Incorporating MV-MoE reduces parameter redundancy and eliminates dual-model deployment, at the cost of a moderate accuracy drop that is consistently observed across configurations due to parameter sharing. On this basis, integrating VL-Rerank further enhances performance by 9.4\% on UAV and 3.2\% on SVI, pushing both tasks beyond the baseline. The full MVLGeo integrates all three components, demonstrating their effectiveness and complementarity.

\begin{table}[!t]
\centering
\small
\setlength{\tabcolsep}{2pt}
\begin{tabular}{cccccc}
\toprule
\makecell{VL-\\Rerank} & \makecell{MV-\\MoE} & \makecell{ESAM-\\Prior} & \makecell{UAV\\Acc@50 (\%)} & \makecell{SVI\\Acc@50 (\%)} & \makecell{Params\\~} \\
\midrule
 $\times$ & $\times$ & $\times$ & 57.66 & 42.24 & 73.8M$\times$2 \\
 $\times$ & $\times$ & $\checkmark$ & 62.58 & 46.76 & 73.8M$\times$2 \\
 $\times$ & $\checkmark$ & $\times$ & 52.14 & 41.35 & 91.0M$\times$1 \\
 $\times$ & $\checkmark$ & $\checkmark$ & 56.63 & 44.09 & 91.0M$\times$1 \\
 $\checkmark$ & $\checkmark$ & $\checkmark$ & \textbf{66.04} & \textbf{47.24} & 91.0M$\times$1 \\
\bottomrule
\end{tabular}
\caption{Ablation study of each component in MVLGeo on the UAV and SVI test sets.}
\label{tab:modular}
\end{table}

\subsubsection{Analysis of MV-MoE Configuration.}

\begin{figure}[!t]
    \centering
    \includegraphics[width=0.85\linewidth]{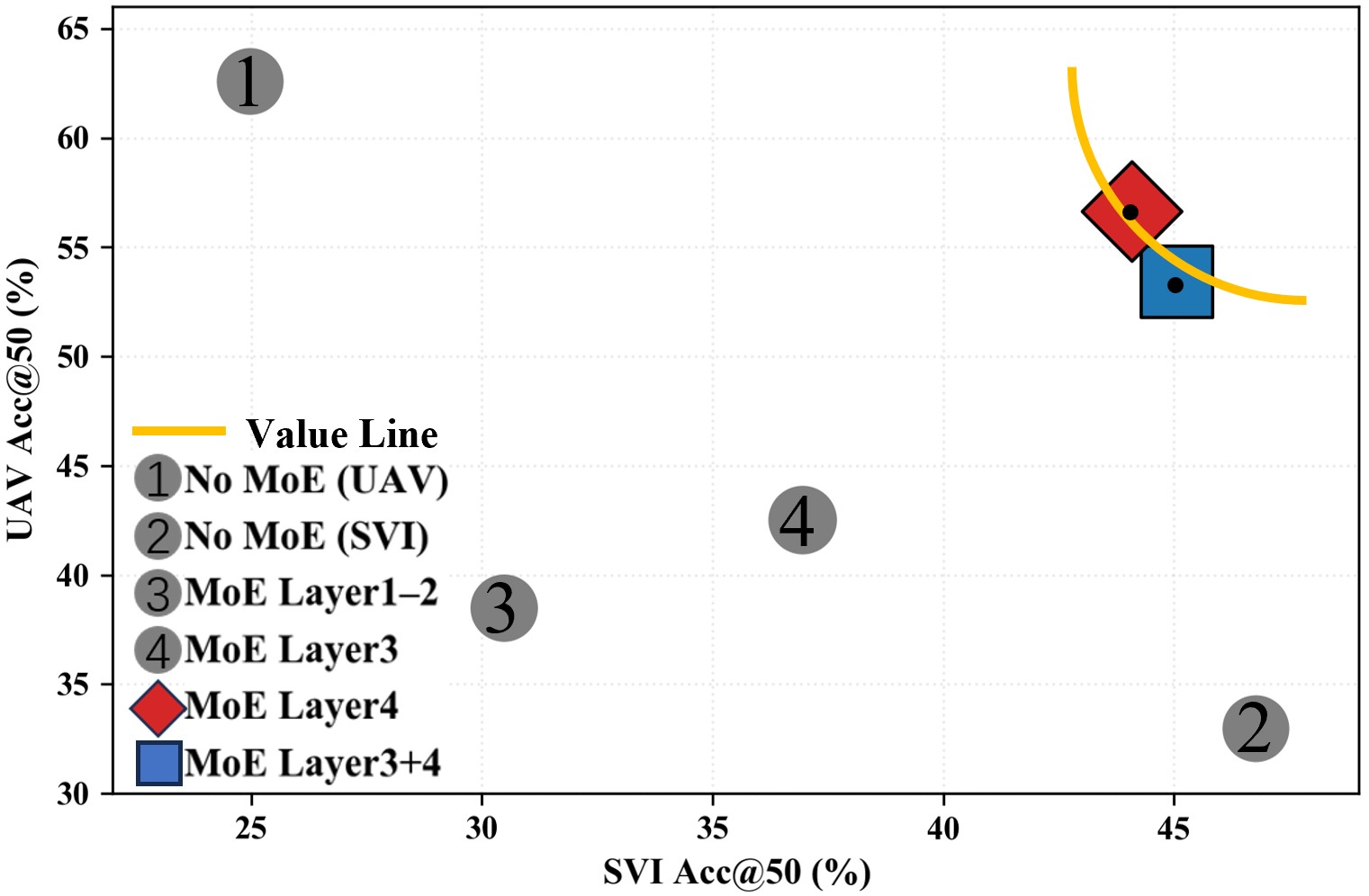}
    \caption{Cross-dataset localization accuracy distribution on UAV and SVI scenes. The yellow value line denotes the Pareto trade-off boundary, and points above the line achieve superior overall performance.}
    \label{fig:ablation_moe_tradeoff}
\end{figure}

We systematically test inserting expert branches at different positions of the feature extraction layers. As shown in Fig.~\ref{fig:ablation_moe_tradeoff}, inserting experts at Layer4 achieves the best cross-view trade-off balance, lying above the Pareto boundary. Layer4 encodes high-level semantics containing prominent cross-view discrepancies, making it suitable for expert modules. By contrast, shallow layers capture view-invariant low-level features, and we observe that inserting experts at these layers yields less than 0.5\% improvement, confirming their limited benefit for domain-specific adaptation. The detection head further calibrates box regression and confidence outputs, complementing backbone feature adaptation.

\begin{table}[!t]
\centering
\small
\setlength{\tabcolsep}{10pt}
\begin{tabular}{ccccc}
\toprule
\# & CL $\lambda$ & BN $\gamma$ init & Val Best & Test Best \\
\midrule
1 & 0.1 & 1.0 (identity) & 49.29 & \textbf{50.36} \\
2 & 0.1 & Random & 44.74 & 43.91 \\
3 & 0.5 & 1.0 & 47.52 & 46.87 \\
4 & 0.0 & 1.0 & \textbf{49.59} & 49.23 \\
\bottomrule
\end{tabular}
\caption{Average Acc@50 (\%) across UAV and SVI on validation and test sets under MV-MoE configurations.}\label{tab:moe_ablation}
\end{table}

We perform ablations on identity initialization and contrastive loss weight, as listed in Tab.~\ref{tab:moe_ablation}. Identity initialization outperforms random initialization by 5.5\% on average. Random initialization injects large gradients from untrained experts through residual connections, destabilizing the pretrained shared backbone from the first iteration. For the contrastive loss, $\lambda=0.1$ achieves the best overall performance. Removing contrastive regularization widens the validation-test gap (overfitting), while excessive regularization imposes overly strict constraints and degrades final accuracy.

\subsubsection{Effects of ESAM-Prior.}
\begin{figure}[!t]
    \centering
    \includegraphics[width=1\linewidth]{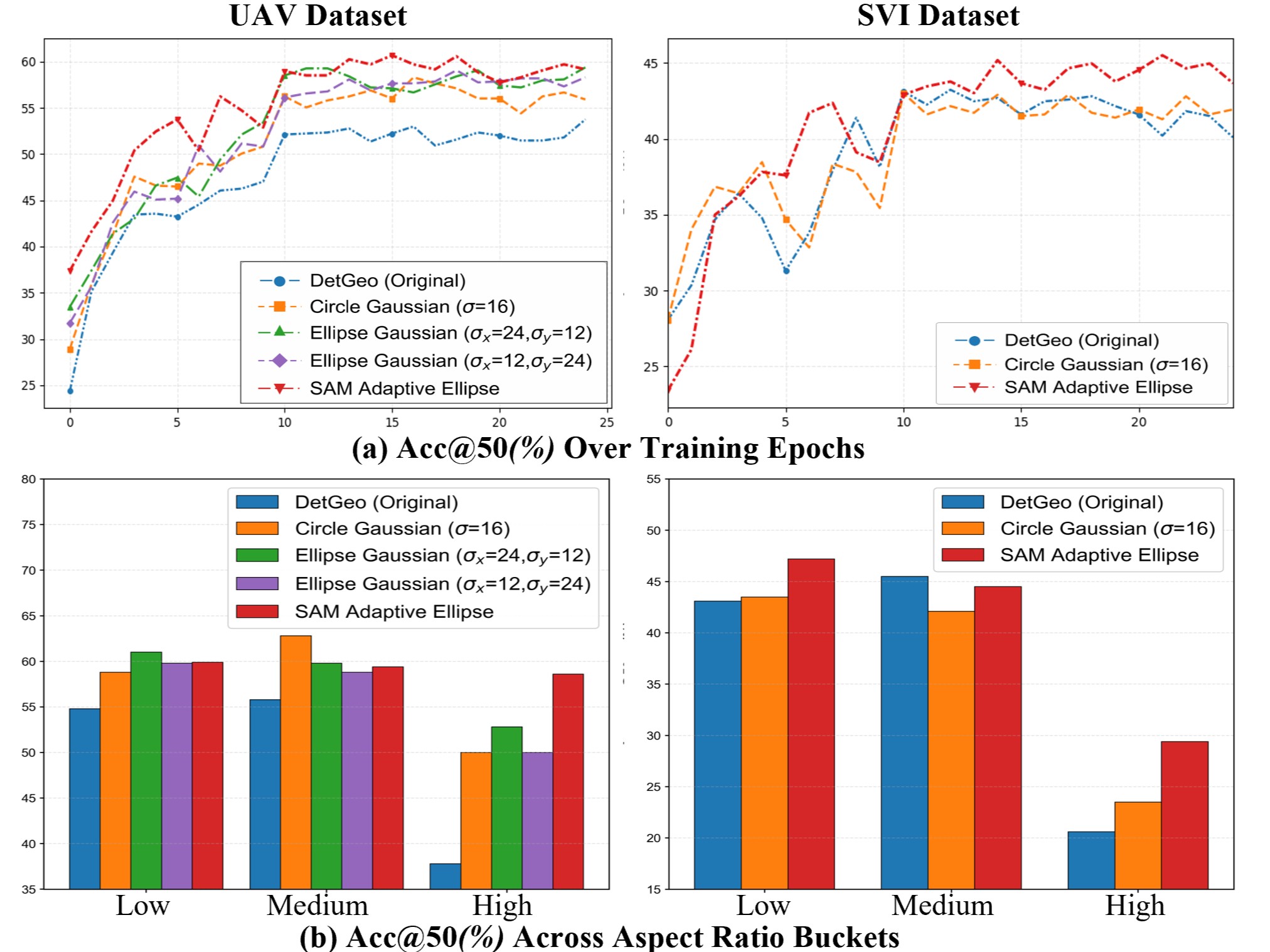}
    \caption{ESAM-Prior ablation results on UAV and SVI datasets. (a) Validation Acc@50 (\%) over training epochs. (b) Acc@50 (\%) across object aspect ratio buckets partitioned into Low, Medium and High groups.}
    \label{fig:Ablation_ESAM-Prior}
\end{figure}
We explore various heatmap encoding strategies from the single-view DetGeo baseline. Full ablation studies are performed on the UAV dataset. Substituting DetGeo's distance map with a fixed isotropic Gaussian yields moderate gains. We further adopt a fixed elliptical prior, improving UAV further and confirming that anisotropic kernels benefit elongated targets. However, a fixed ellipse cannot adapt to varying instance shapes. Inspired by this, we propose ESAM-Prior to produce instance-adaptive elliptical priors, achieving the optimal result on UAV. We then verify generalization on the SVI dataset with three representative schemes, including DetGeo, fixed circular Gaussian, and ESAM-Prior, and consistent gains are obtained.

As shown in Fig.~\ref{fig:Ablation_ESAM-Prior}, ESAM-Prior consistently outperforms baselines under all conditions. Compared to isotropic priors that spread activation into background regions for elongated objects, ESAM-Prior produces tighter heatmaps along the target's major axis, as shown in Fig.~\ref{fig:ESAM-Prior}. Specifically, improvements are most prominent for elongated targets with aspect ratio of 3.0 or higher. In the High bucket, ESAM-Prior achieves the largest gains. Medium-bucket samples show marginal differences among all methods.

\begin{table}[!t]
\centering
\small
\setlength{\tabcolsep}{3pt}
\begin{tabular}{lccc}
\toprule
Test modality & Single baseline & MV-MoE unified & Retention \\
\midrule
UAV Vis (UV) & 62.6 & \textbf{56.6} & 90.4\% \\
UAV Gray (UG) & 62.4 & \textbf{54.5} & 87.3\% \\
SVI Vis (SV) & 46.8 & \textbf{44.0} & 94.0\% \\
SVI Gray (SG) & 45.9 & \textbf{44.1} & 96.1\% \\
\midrule
\textbf{4-View Avg} & --- & \textbf{49.8} & --- \\
\bottomrule
\end{tabular}
\caption{Unified MV-MoE generalizes across four modality-viewpoint combinations with minimal accuracy degradation under grayscale inputs.}\label{tab:unified}
\end{table}

\subsection{Extension to Other Input Modalities}

The parameter-sharing property of MV-MoE generalizes effectively to grayscale inputs without retraining, as shown in Tab.~\ref{tab:unified}. By training on visible-light data and applying the same unified model directly to grayscale at test time, MV-MoE retains competitive accuracy across all four modality-viewpoint combinations with minimal degradation. These results further demonstrate the broad modality robustness of the proposed unified architecture.

\section{Conclusion}

In this paper, we presented MVLGeo, a unified multi-view framework for cross-view object geo-localization. MVLGeo addresses parameter redundancy through shared-residual Mixture-of-Experts, resolves candidate ambiguity via vision-language reranking with environmental context, and provides shape-aware spatial encoding through instance-adaptive elliptical priors. Extensive experiments demonstrate that MVLGeo surpasses existing dual-detector methods with a single unified model and generalizes robustly to unseen modalities without retraining. These results open up new possibilities for scalable multi-view deployment in geo-localization.

\textbf{Limitations.}
The MVLGeo architecture has not been extended beyond the CVOGL benchmark. Environmental descriptions generated from street-view queries suffer from perspective distortion, reducing the reranking gain on this task. Prompt engineering remains manual.

{\small
\bibliographystyle{ieee_fullname}
\bibliography{references}
}

\end{document}